\documentclass[10pt,twocolumn,letterpaper]{article}

\usepackage{cvpr}              

\usepackage[dvipsnames]{xcolor}

\definecolor{cvprblue}{rgb}{0.21,0.49,0.74}
\usepackage[pagebackref,breaklinks,colorlinks,citecolor=cvprblue]{hyperref}

\usepackage{amsthm}
\newtheorem{theorem}{Theorem}[section]

\theoremstyle{definition}
\newtheorem{definition}[theorem]{Definition}

\theoremstyle{remark}

\def\paperID{*****} 
\def\confName{CVPR}
\def\confYear{2024}

\title{RITA: A \underline{R}eal-time \underline{I}nteractive \underline{T}alking \underline{A}vatars Framework}

\author{Wuxinlin Cheng\thanks{These authors contributed equally to this work.}\\
Stevens Institute of Technology\\
{\tt\small wcheng7@stevens.edu}
\and
Cheng Wan\footnotemark[1]\\
Georgia Institute of Technology\\
{\tt\small cwan38@gatech.edu}
\and
Yupeng Cao\footnotemark[1]\\
Stevens Institute of Technology\\
{\tt\small ycao33@stevens.edu}
\and
Sihan Chen\footnotemark[1]\\
University of Illinois Urbana-Champaign\\
{\tt\small sihan6@illinois.edu}
}

\begin{document}
\maketitle
\begin{abstract}
RITA presents a high-quality real-time interactive framework built upon generative models, designed with practical applications in mind. Our framework enables the transformation of user-uploaded photos into digital avatars that can engage in real-time dialogue interactions. By leveraging the latest advancements in generative modeling, we have developed a versatile platform that not only enhances the user experience through dynamic conversational avatars but also opens new avenues for applications in virtual reality, online education, and interactive gaming. This work showcases the potential of integrating computer vision and natural language processing technologies to create immersive and interactive digital personas, pushing the boundaries of how we interact with digital content.

\end{abstract}    
\section{Introduction}
\label{sec:intro}
Animating a single image to create dynamic speech-driven facial animations lies at the intersection of artificial intelligence, computer vision, and multimedia technology. The advent of generative models has significantly advanced the creation of talking head videos, where a static image is animated to match spoken audio, transforming a once stationary depiction into a lively representation of speech. However, this field faces challenges due to the inherent delay in video generation, as transforming an image and audio into a seamless video sequence requires significant computational resources and time, often making real-time applications challenging to achieve.

Drawing inspiration from state-of-the-art models such as SadTalker and MakeItTalk~\cite{sadtalker, makeittalk}, which pioneered in generating lip-sync videos through sophisticated facial modeling and motion synthesis. Our Real-time Interactive Talking Avatar (RITA) framework transcends the limitations of its former generation speed and unsustainable interactions. While SadTalker and similar models excel in their domain, they are primarily tethered to offline processing, owing to the intricate computations required to ensure synchronicity between audio cues and facial movements, including lip motion, head pose, and eye blinks. These models, despite their efficacy, fall short in applications demanding real-time interaction, thus limiting their utility in dynamic, user-centric scenarios.

RITA emerges as a groundbreaking framework, engineered to bridge this gap by facilitating an end-to-end real-time dialogue solution. At its core, RITA leverages a novel architecture that combines the immediacy of real-time processing with the depth of generative models to animate static portraits. By integrating a live-feed mechanism, users can interact with avatars in a seamless, responsive environment, where avatars not only lip-sync but also exhibit natural head movements and expressions, closely mimicking human interactions. This leap in technological innovation is made possible through the strategic application of lightweight models, optimized for speed without compromising the richness of the avatar's responsiveness.

Moreover, RITA introduces an innovative application of Large Language Models (LLMs) for content generation, allowing avatars to engage in coherent, contextually relevant dialogues. This integration not only enhances the interactive experience but also expands the potential applications of talking avatars, ranging from virtual customer service agents to personalized digital companions.

In RITA, we refined the generative process to ensure high-definition output and fluency in avatar-user interactions, setting a new benchmark for real-time performance in talking avatar technology. Our contributions are manifold, addressing both the technical challenges in achieving real-time interactivity and the practical implications of deploying such technology in user-centric applications.

The following sections will delineate the architecture of RITA, the innovations in real-time processing, and the synergistic use of LLMs for dynamic content generation. Through comparative analysis and empirical validation, we will demonstrate RITA's superiority over existing models in terms of latency, quality, and applicability, thereby heralding a new era in interactive digital avatars. In summary, our contributions include:
\begin{itemize}
    \item 
    The development of RITA, an end-to-end framework for generating real-time interactive talking avatars from static images, leveraging advancements in generative models and LLMs for naturalistic avatar-user dialogues.
    \item 
    A novel real-time processing pipeline that significantly reduces the latency observed in previous talking head models, enabling seamless, high-fidelity interactions.
    \item 
    The integration of LLMs for content generation, providing a versatile platform for avatars to engage in context-aware conversations with users.
    \item 
    Empirical evidence showcasing the enhanced performance of RITA over existing methods, highlighting improvements in speed, interaction quality, and user engagement.
\end{itemize}

\section{Related Work}
\label{sec:background}

\subsection{Audio-driven Talking Head Generation}
\label{subsec:talkinghead}

The first batch of Audio-driven Talking Head Generation work focused on building videos by generating lip movements from audio inputs, such as Synthesizing Obama~\cite{suwajanakorn2017synthesizing}, LMGG~\cite{chen2018lip}, Wav2Lip~\cite{prajwal2020lip}, and TalkLip~\cite{wang2023seeing}. While these methods effectively synchronized lip movements with audio, they often overlooked other facial and head movement details, limiting the realism of the generated heads.

To address these limitations, Audio-Visual Mapping approaches emerged, focusing on a more comprehensive representation of facial details. AD-AVR~\cite{zhou2019talking} employed an adversarial learning framework to separate audio and visual representations, allowing independent control over facial expressions and lip movements. The separated representations can independently control facial expressions and lip movements. ATVGNet~\cite{chen2019hierarchical} introduced a hierarchical cross-modal learning framework, enhancing facial expressions and lip motion through integrated audio-visual information.  MakeItTalk~\cite{makeittalk} addressed speaker-specific characteristics by separately handling speaker and lip motion generation, while Audio2Head~\cite{wang2021audio2head} utilized a motion-aware recurrent neural network for predicting head movements. Despite these advancements, challenges such as producing vivid head movements and maintaining accurate facial identities persisted.

The field then shifted towards methods inspired by video editing, particularly those using 3DMM information for face reconstruction and animation. AD-NeRF~\cite{guo2021ad} and subsequent works like \cite{zhang2021facial} and \cite{ji2021audio} demonstrated significant improvements in mouth shape editing, head movement, and emotional expressiveness. However, these methods faced limitations in generalizing to arbitrary photos and audio input. Addressing this, SadTalker~\cite{sadtalker} introduced a novel approach by modulating 3D perceptual face rendering with 3D motion coefficients derived from audio. SadTalker showed promise but did not fully cater to real-time applications in diverse real-world scenarios.

\subsection{Large Language Model for Vision Applications}
\label{subsec:llm}

The emergence of powerful LLMs like ChatGPT~\cite{chatgpt} and Llama~\cite{llama} has revolutionized natural language understanding and dialogue generation with human-like responsiveness and sophisticated context-generation capabilities~\cite{chang2023survey, safdari2023personality, xu2023exploring}. Benefiting from LLM development, the trend of using autoregressive language models as decoders in vision-language tasks has gained significant traction~\cite{huang2024language, li2023blip, driess2023palm, jia2023kg}. A substantial body of work demonstrates the efficacy of merging visual features with language models to enhance reasoning capabilities in complex vision-language tasks. For instance, LENS~\cite{berrios2023towards} employs a synergy of raw visual data and a frozen LLM for image captioning tasks. By combining LLM with the underlying vision model, the door can be opened to open-world understanding, reasoning, and a small amount of learning that is lacking in current autonomous driving systems~\cite{yang2023survey}. However, the application of LLMs in talking head generation, a field requiring nuanced integration of audio-visual and linguistic elements, remains a largely unexplored frontier.

Inspired by this gap and the recent strides in Talking Head Generation, we use LLMs to build a dialog system in the RITA framework, aiming to optimize real-time performance while improving the user experience.

\section{RITA Design}
\begin{figure*}[ht]
\centering
  \includegraphics[width=1\linewidth, trim={0cm 7cm 0cm 2.5cm}, clip]{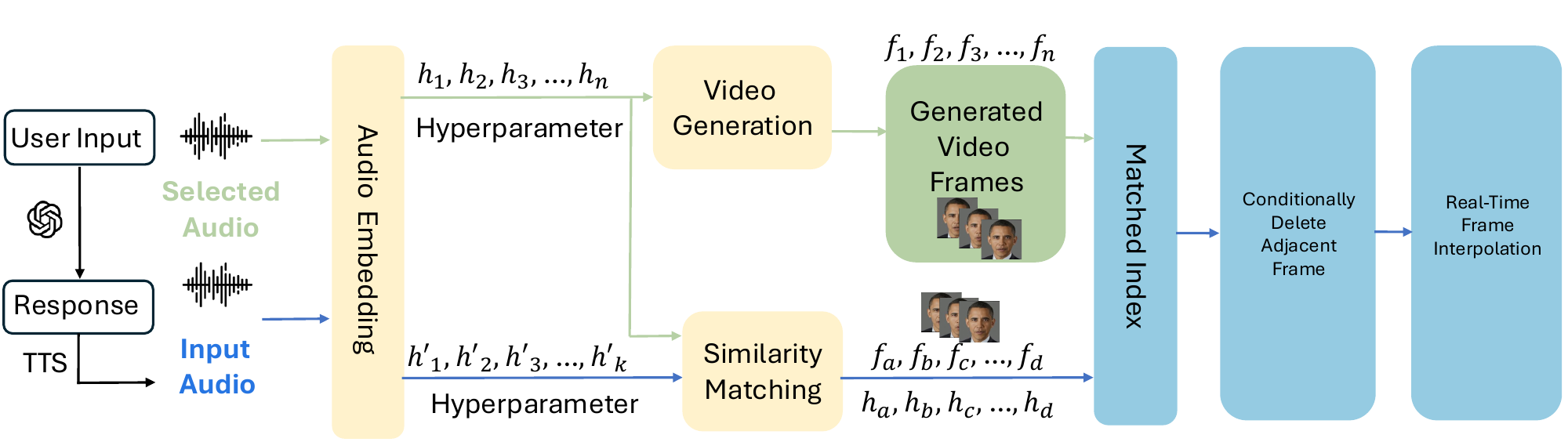}
\caption{The overview of RITA. Notice green arrows indicate foundational frame generation, which is not required in real-time inference. The real-time inference only requires the bottom blue arrows.}
\label{fig:pipeline}
\end{figure*}

Figure~\ref{fig:pipeline} is the overview of the RITA framework. RITA is a framework that introduces a transformative approach to generating talking avatars in real-time from a single image. This section explores the intricacies of RITA, detailing the innovative components and processes that enable this advanced capability.

\subsection{Phase 1: Foundational Frames Generation}
User-entered conversations are first fed into LLMs to generate personalized responses, which are then converted to audio format by text-to-speech (TTS) model. RITA will perform video generation based on the converted audio and the selected audio. 

The initial phase of RITA commences with the generation of foundational frames (videos). This process utilizes a flexible approach, allowing for the incorporation of many open-source audio-driven talking-head model~\cite{sun2023vividtalk,zhang2023sadtalker,wang2021audio2head,gan2023efficient}. These audio-driven models can be bifurcated into two sequential components: the first segment is responsible for the embedding of auditory signals into a set of hyperparameters $H\in \mathbb{R}^{|K| \times |N|}$, effectively capturing the essence of the speech in a structured form; the subsequent segment leverages these hyperparameters to generate the corresponding visual frames $F\in \mathbb{R}^{|K|}$. The primary objective of this phase is to synchronize generated frames with their respective hyperparameters, a process pivotal for the seamless progression to subsequent phases. To generate the desired frames, we strategically employ a collection of a few audio samples, meticulously selected to encompass a wide spectrum of lip movements. This selection aims to augment the avatar's expressive capacity, ensuring a diverse and naturalistic simulation of speech articulation. 

To balance the dual imperatives of quality and efficiency, each frame generated during this preliminary phase is saved in a high-quality compressed image format, such as JPEG. This approach not only preserves the fidelity of the visual output but also optimizes storage. At the same time, the generated frames form a comprehensive library of lip movements that RITA will also utilize in the subsequent phase.

\subsection{Phase 2: Dynamic Frame Matching}
Building upon the initial generation from the last phase, each frame $F$ of the foundational video is precisely annotated with hyperparameters $H$.

When RITA is presented with new audio inputs, it employs the first segment of the talking-head model, as mentioned in the preceding phase. This critical operation transforms the new audio into a set of hyperparameters, denoted as $H' \in \mathbb{R}^{|X| \times |N|}$, where $X$ represents the dimensionality of the input space and $N$ signifies the number of hyperparameters. This enables RITA to swiftly adapt to new audio stimuli, circumventing the computationally intensive task of generating entirely new frames. The subsequent step involves a meticulous matching process, where each newly embedded hyperparameter $H'$ is compared against the pre-generated set of hyperparameters $H$. The goal of this comparison is to identify the most congruent pair, thereby selecting the initially generated frame $F$ that best matches the new audio input. To facilitate real-time comparison of a vast array of frames, a high-performance algorithm can be deployed, such as the approximate nearest neighbor~\cite{arya1998optimal}. Thereby, RITA circumvents the need for continuous video regeneration.

Moreover, the regenerated video can be further refined through an intelligent frame reduction strategy. By methodically comparing hyperparameter similarity distances between sequential frames.

\begin{definition}
For predicted hyperparameter $H'_{i} \in \mathbb{R}^{|N|}$ and its matching hyperparameter$H_{j} \in \mathbb{R}^{|N|}$. The similarity distance is defined as $SD_{i} = \sum_{n=1}^{|N|} \frac{|H_{in} - H'_{jn}|}{\max(H_{in}, H'_{jn})}$
\end{definition}

For every two sequential frames $f_t$ and $f_{t+1}$ and their corresponding $SD_t$ and $SD_{t+1}. $RITA identifies and eliminates the frame with the larger similarity distances. This selective reduction process effectively maintains the frame's visual coherence.

\subsection{Phase 3: Real-Time Video Interpolation}
Despite the strategic frame reduction in the preceding phase, which effectively halved the video's frame rate, a notable challenge arises: the resulting video still lacks the desired smoothness, exhibiting noticeable jumps and transitions between frames. To address this and restore the natural fluidity of the avatar's movements, RITA employs advanced Real-Time Video Frame Interpolation techniques at this crucial juncture, such as RIFE~\cite{huang2022real}

The essence of this phase lies in its ability to intricately fill the gaps left by the frame reduction strategy. By interpolating new frames between the existing ones, this technique not only increases the frame rate back to its original or even a higher level but also significantly smooths out the transitions, ensuring that the avatar's expressions and movements regain their lifelike flow. This process is pivotal for enhancing the overall realism of the generated video, making the talking avatar's movements appear seamless and naturally aligned with the audio input.

\section{Application and Performance Evaluation}
This section presents a comprehensive evaluation of the RITA framework, showcasing its versatility and efficiency across different operational scenarios. We detail the results stemming from varying numbers of foundational frames, the impact of diverse image inputs on the final avatar animations, the runtime of each critical phase within the framework, and the specifications of the hardware platform utilized for our experiments.

\subsection{Foundational Frame Generation and File Size}
\begin{figure}[ht]
\centering
  \includegraphics[width=1\columnwidth, trim={9.3cm 0 9.5cm 0}, clip]{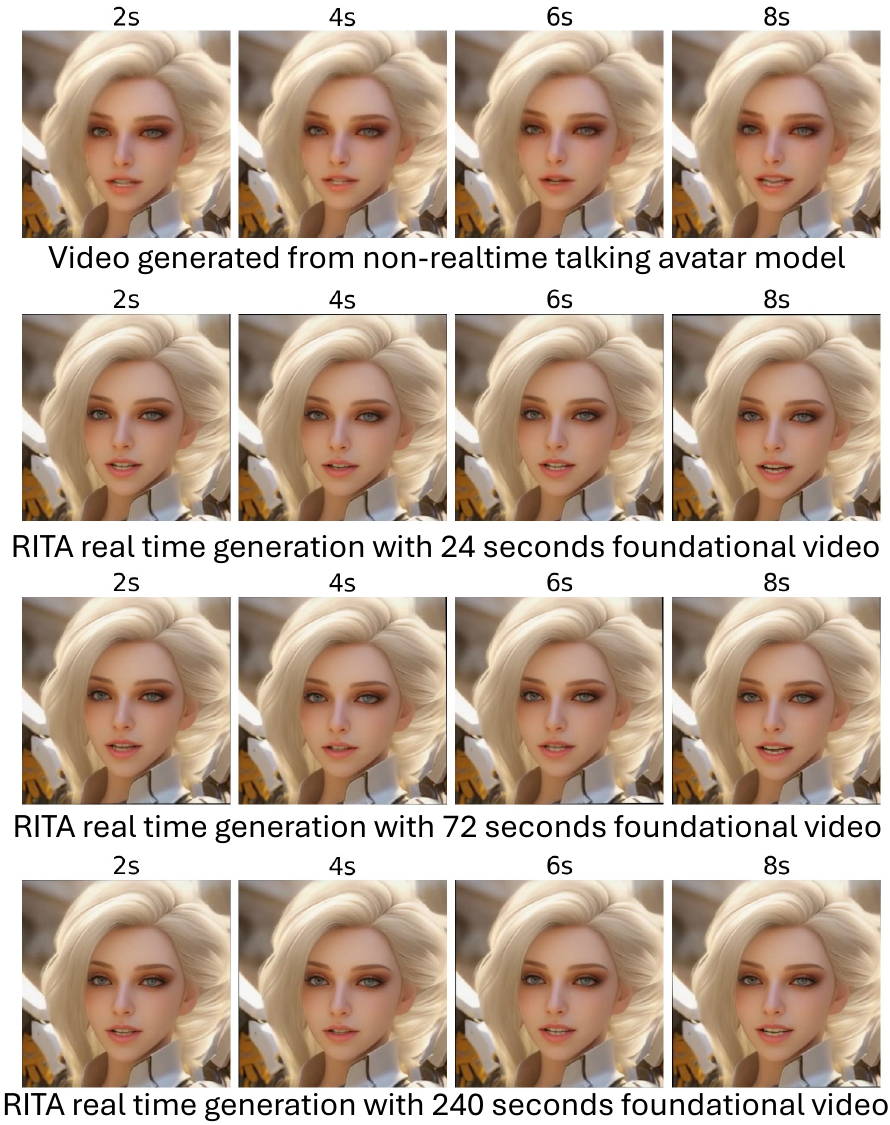}
\caption{Frame comparison between non-realtime avatar generative model and RITA. Keyframe times are 2, 4, 6, and 8 seconds.}
\label{fig:overwatch}
\end{figure}

We initiated our analysis by generating foundational videos with the RITA framework. various distinct audio samples are utilized to cover a wide spectrum of lip movements. To balance the trade-off between animation smoothness and storage, we explore varying numbers of foundational videos as Figure~\ref{fig:overwatch}. All videos are saved in JPEG format to save storage space and RITA performs all generation in real-time.

\begin{itemize}
    \item \href{https://drive.google.com/file/d/1FElFKEiaiToT7Yd92wIlXqcKNx6N-cp_/view?usp=sharing}{Example Video 1}: RITA generated video 1 with a 24 seconds foundational video (around 30MB).
    \item \href{https://drive.google.com/file/d/1mHgkrMpR1eyv7xiU5AmVtpYgY_14dveM/view?usp=sharing}{Example Video 2}: RITA generated video 2 with a 72 seconds foundational video (around 90MB).
    \item \href{https://drive.google.com/file/d/1XLyz1sGFdBPkj1OppYPdV5g0FtOy1lfz/view?usp=sharing}{Example Video 3}: RITA generated video 3 with a 240 seconds foundational video (around 300MB).
\end{itemize}

These variations demonstrate the scalable nature of RITA, allowing for customization based on the desired balance between visual fidelity and file size.

\subsection{Impact of Different Image Inputs}
The framework's adaptability was further tested with different image inputs, ranging from stylized 2D animations to photorealistic images of real humans. The goal was to evaluate RITA's robustness and the quality of the final talking avatar across varied input conditions.

\begin{table}[h]
\centering
\caption{\href{https://drive.google.com/file/d/15T2YfpRhrWoSaqBVEi8eDcjvPpqhD0py/view}{Visual progression of avatars}}
\label{tab:image_progression}
\begin{tabular}{c c c c}
\hline
\textbf{Avatars} & \textbf{Initial} & \textbf{Midway} & \textbf{Final} \\ \hline
Real Humans & \includegraphics[width=0.5in]{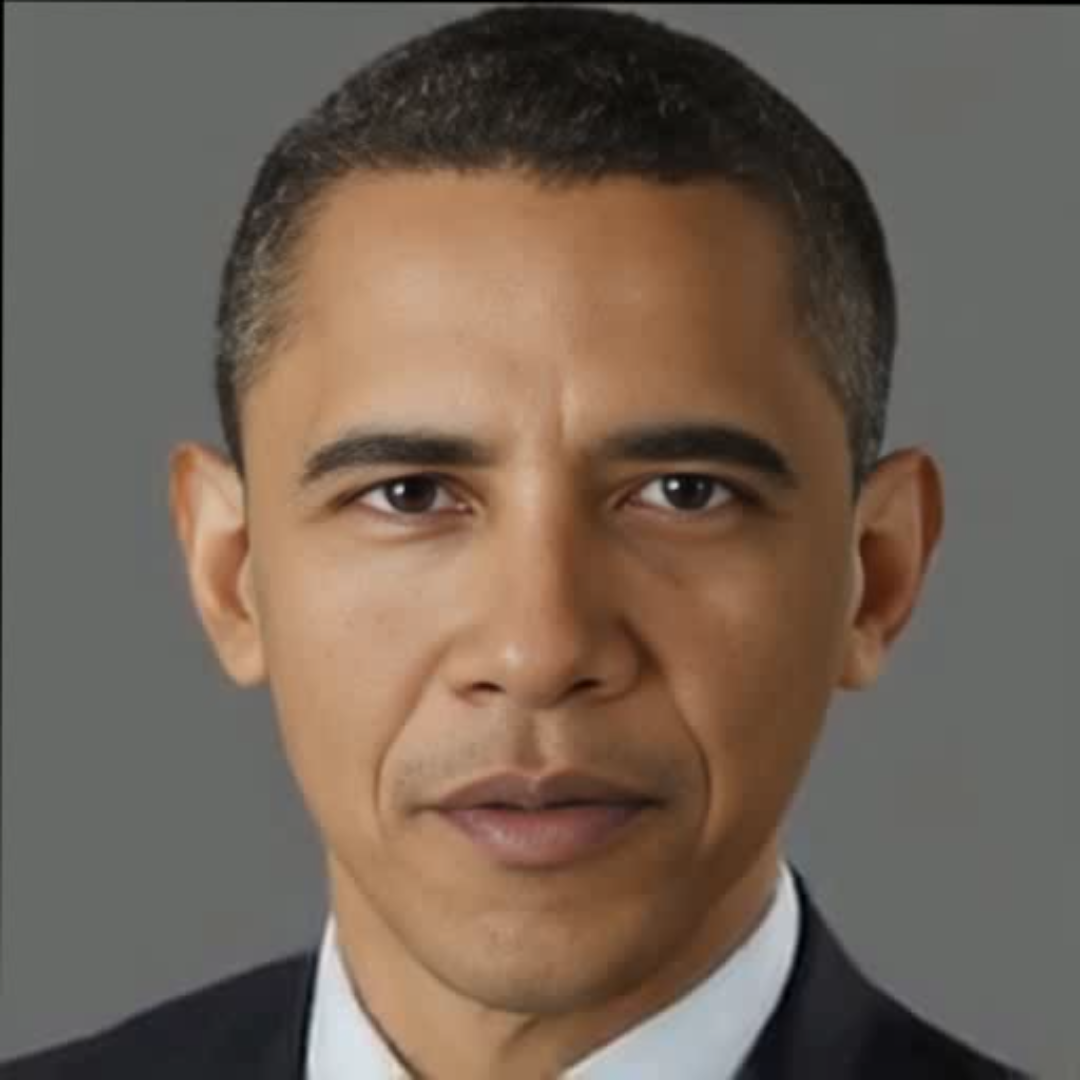} & \includegraphics[width=0.5in]{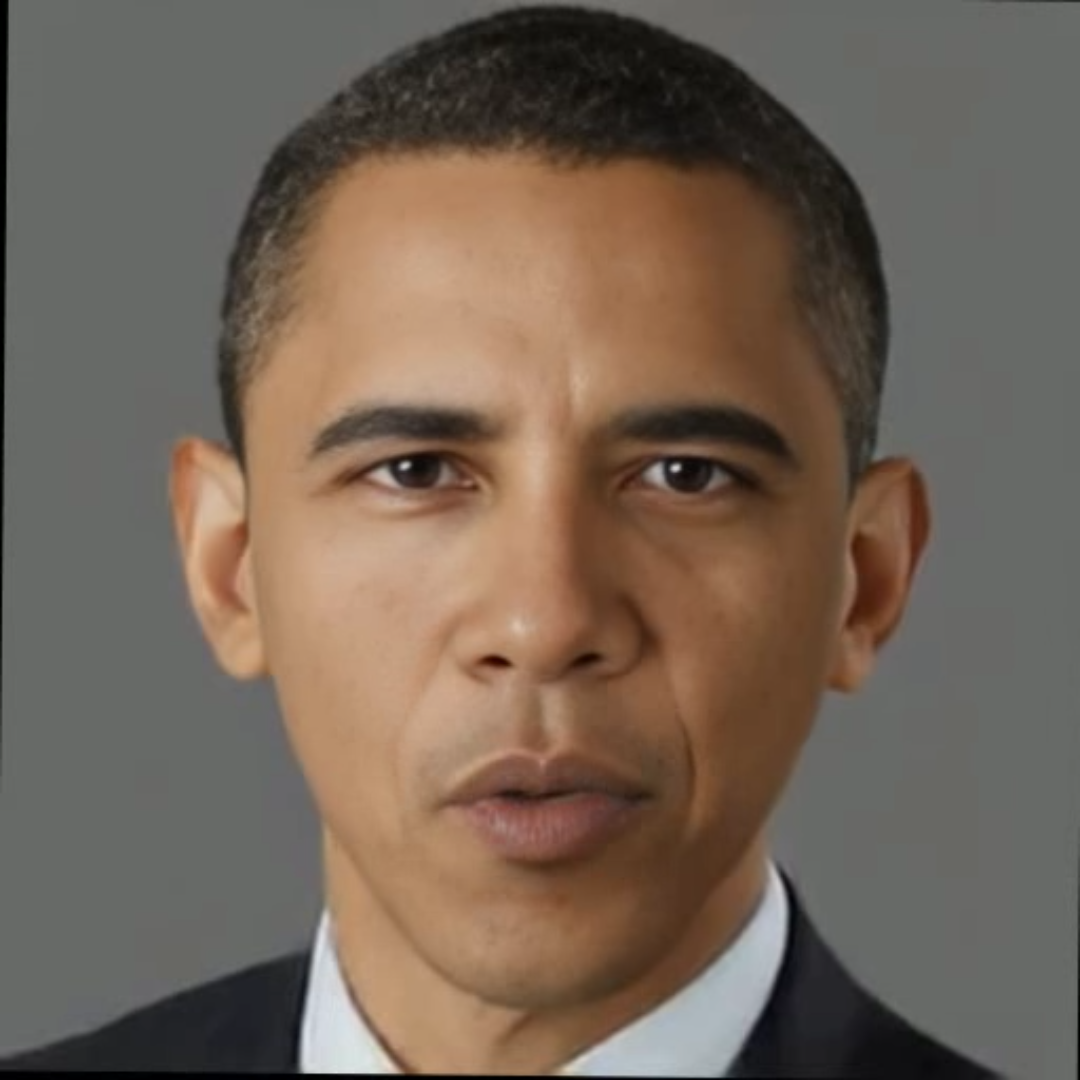} & \includegraphics[width=0.5in]{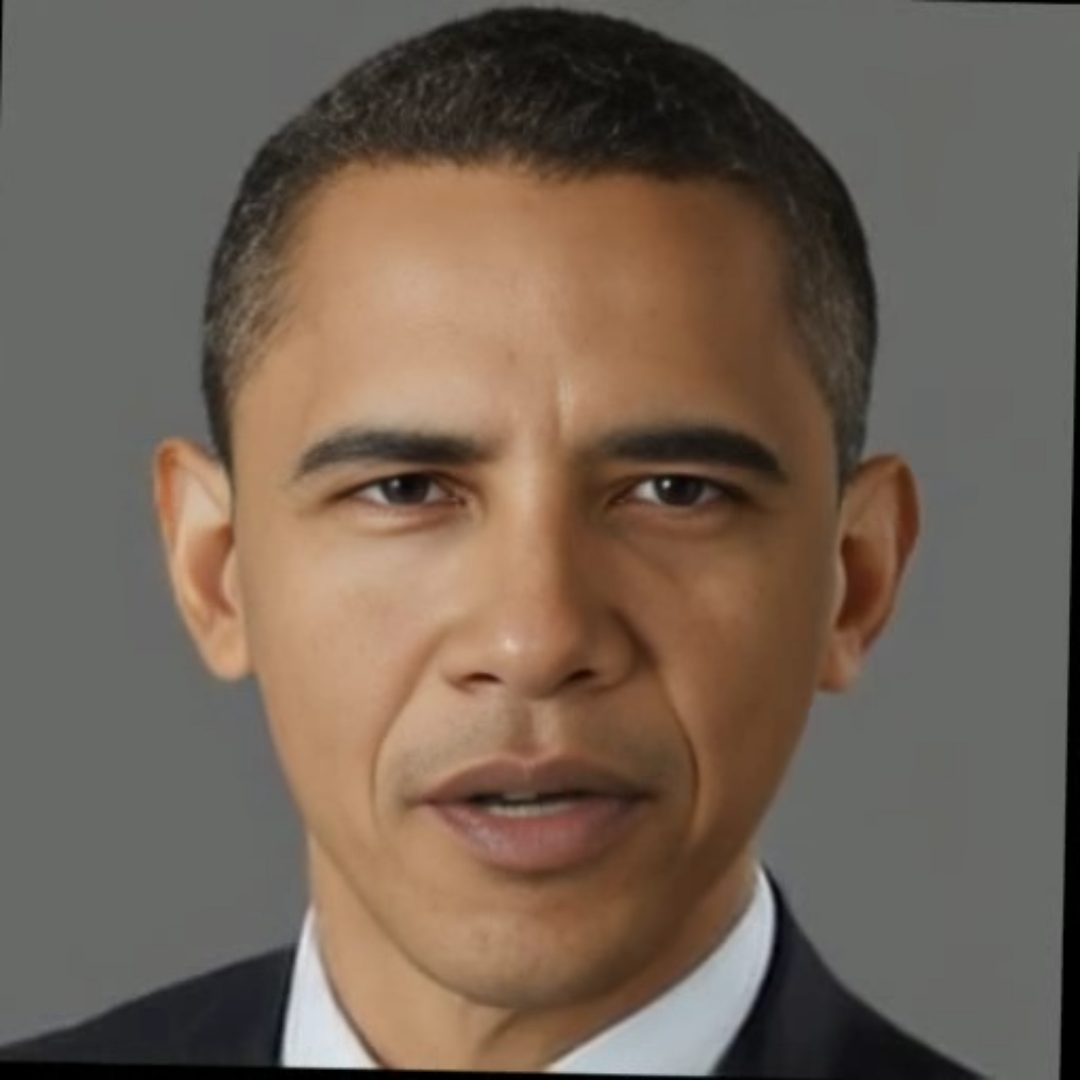} \\
Anime Avatars & \includegraphics[width=0.5in]{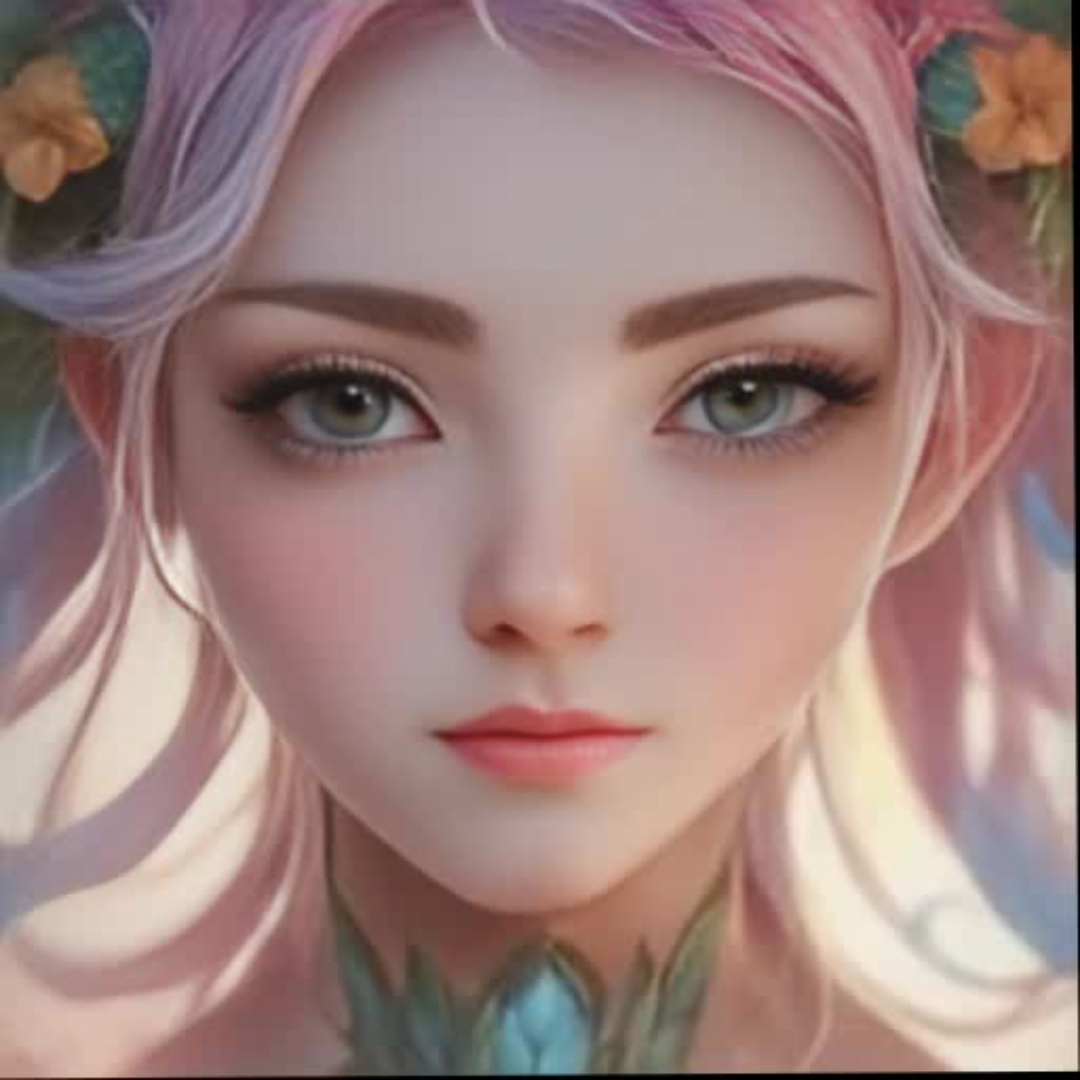} & \includegraphics[width=0.5in]{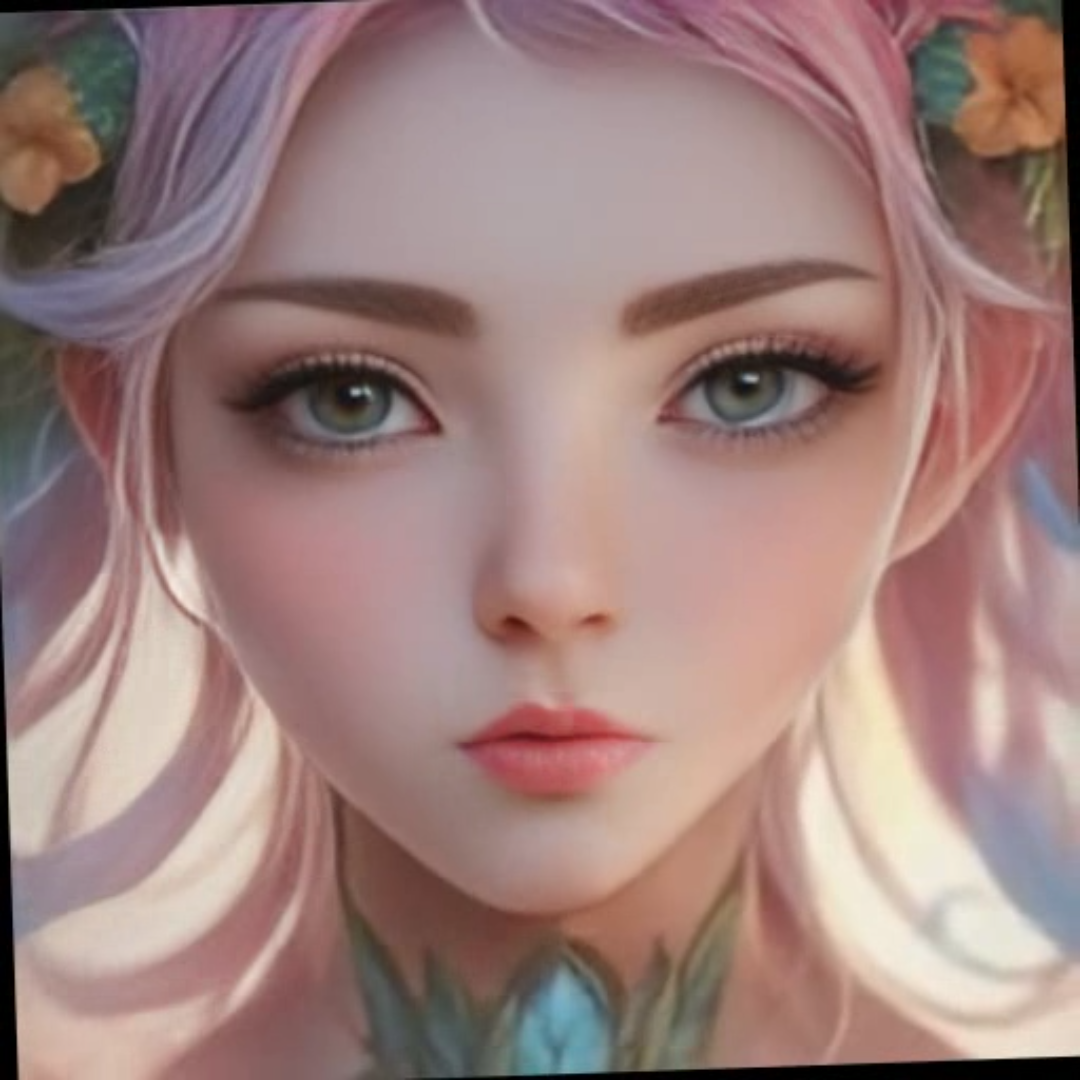} & \includegraphics[width=0.5in]{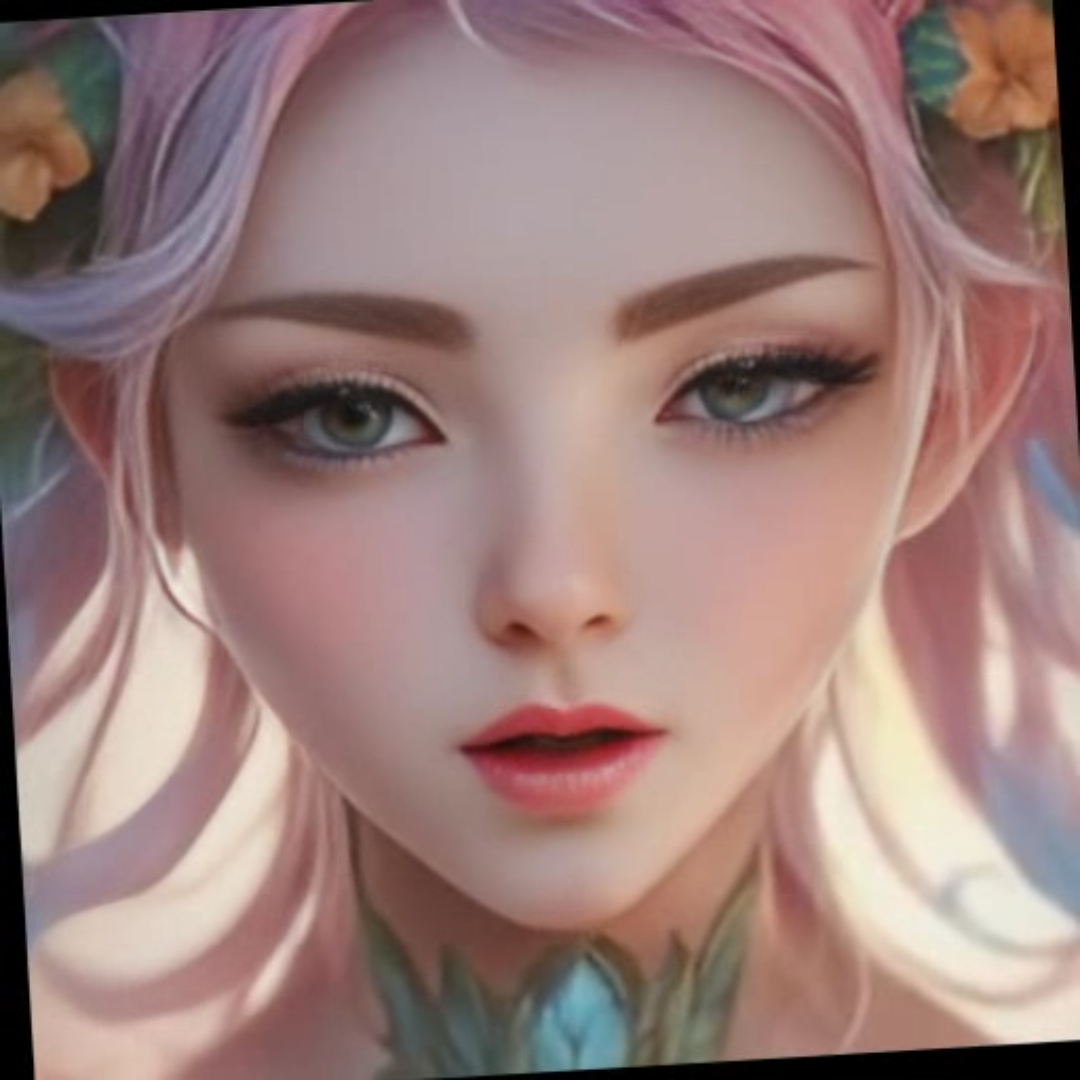} \\
Portrait Painting & \includegraphics[width=0.5in]{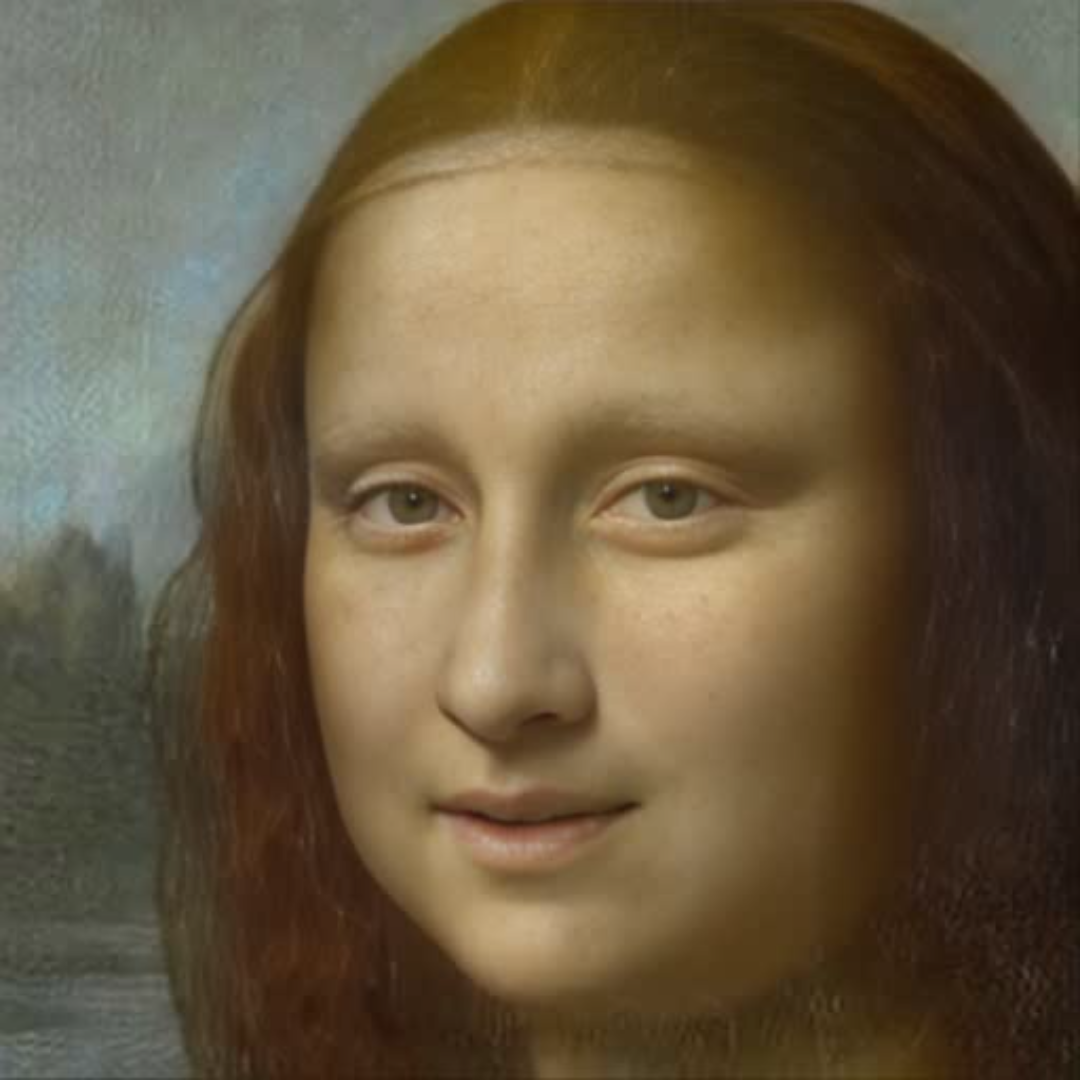} & \includegraphics[width=0.5in]{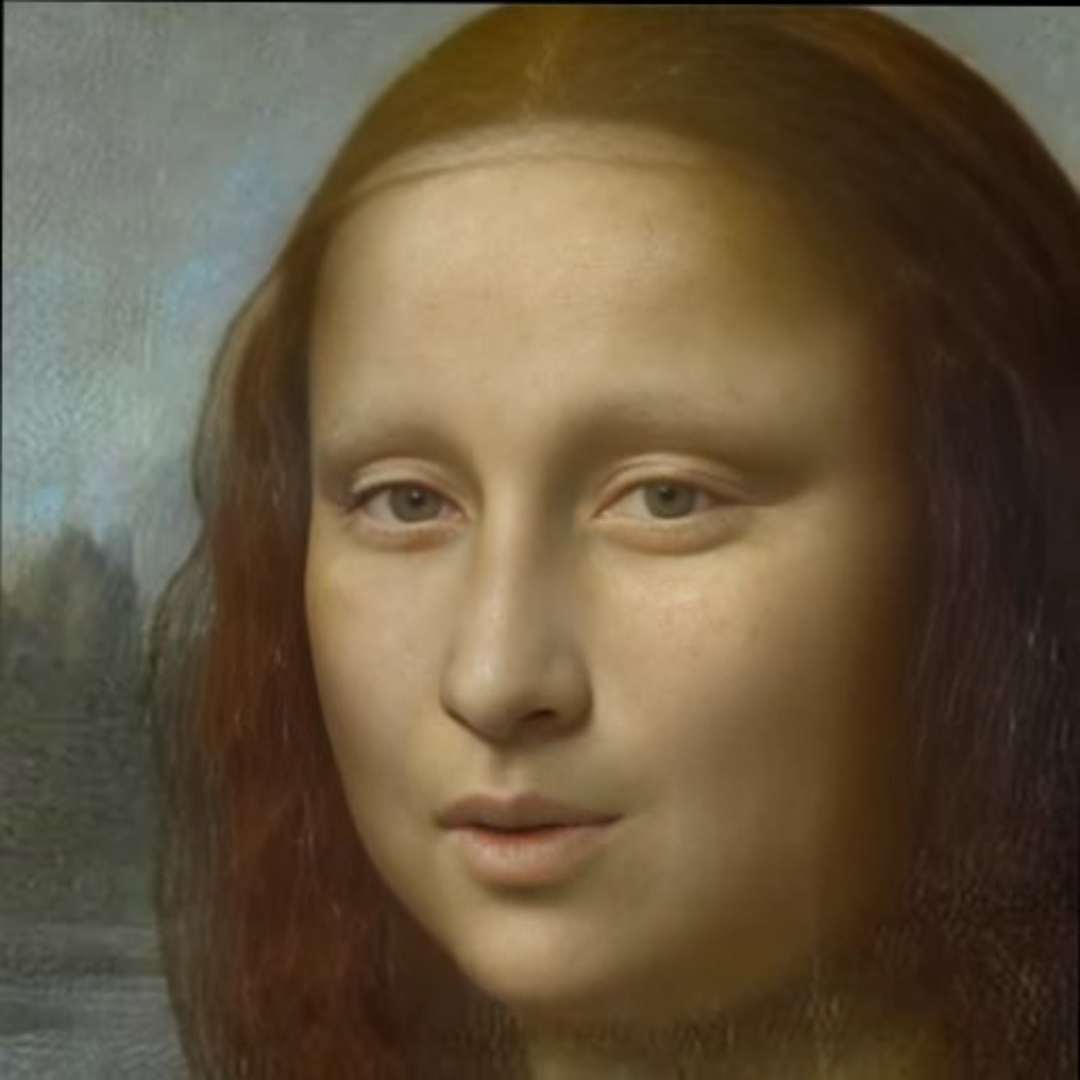} & \includegraphics[width=0.5in]{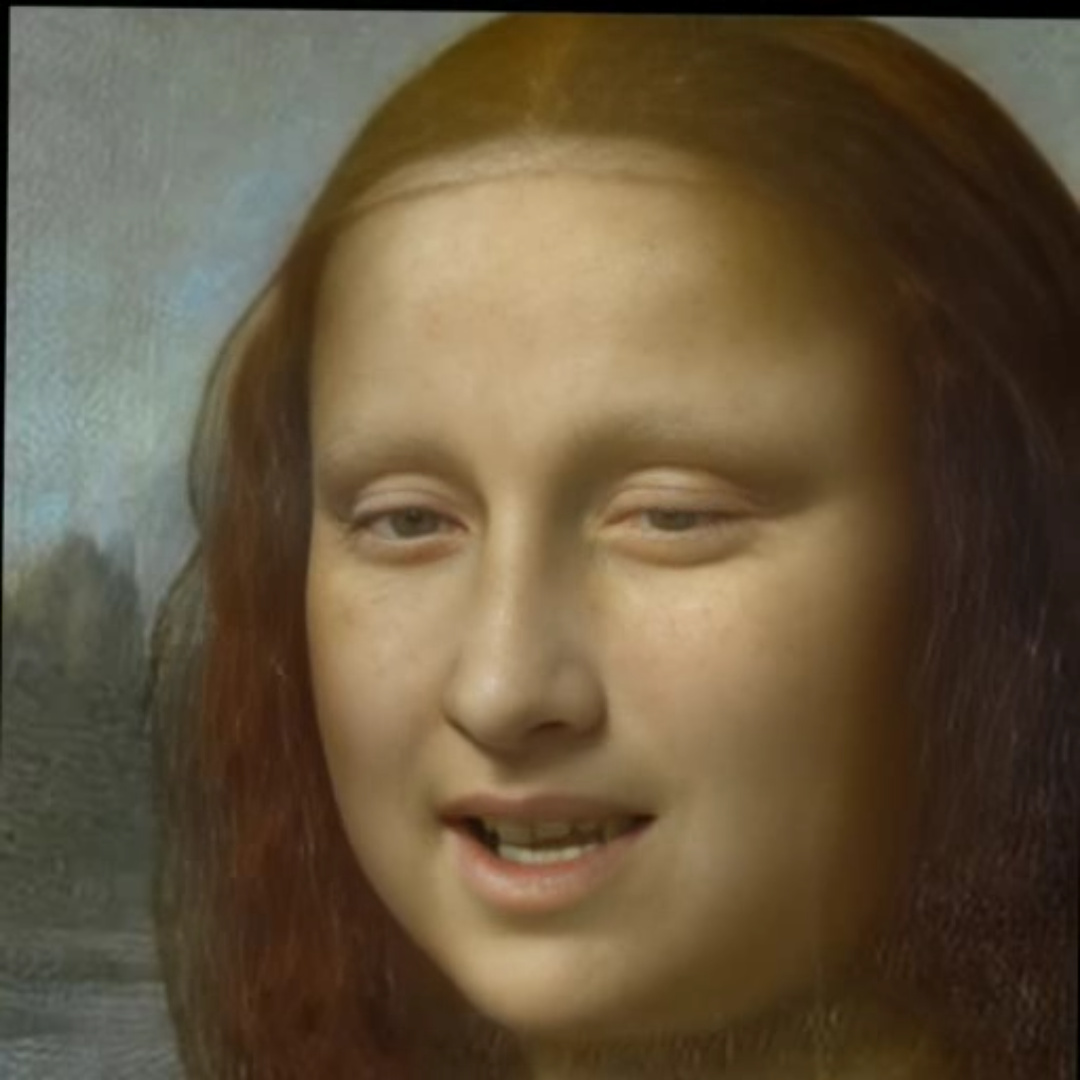} \\
\hline
\end{tabular}
\end{table}

The framework has shown an exceptional capability in handling various avatars with remarkable lip synchronization and fluid facial animations, notably in eye movements. 

\begin{itemize}
    \item \textbf{2D Animation Characters}: The framework adeptly handled the stylized features and exaggerated expressions typical of 2D animations, producing engaging and expressive talking avatars that retained the charm and characteristics of the original artwork.

    \item \textbf{Real Human Images}: When applied to photorealistic images of real humans, RITA demonstrated remarkable precision in lip synchronization and facial animation, creating lifelike avatars that closely mimicked the nuances of human speech and expressions.

    \item \textbf{Portrait Painting}: The adaptation of classic artworks like the Mona Lisa into talking avatars presented best facial movement. The framework managed to animate these paintings with a surprising degree of liveliness, particularly in the eye movements, which are crucial for conveying emotion and intent.
\end{itemize}

These findings underscore the framework's flexibility and its capacity to maintain performance across different types of image inputs, achieving a high level of realism and engagement, particularly in the subtle but crucial area of eye and lip movements.

\subsection{Runtime Analysis}
The efficiency of RITA was quantitatively assessed by measuring the runtime of each phase on a standard hardware setup. The analysis highlighted the real-time capability of the framework, crucial for interactive applications.

\begin{itemize}
    \item Hyperparameter Embedding and Frame Matching: After the input audio is embeded into parameters, the similarity is matched with the parameters of the generated audios corresponding frames, and a series of frames that best match the input audio are output. Therefore, the audio is embedded as a parameter matrix and then matched with the generated existing parameter library to have a sequence of initial video frames. In this stage, we performed 10 average calculations and obtained the average running and calculation time required to be \textbf{0.09} seconds which is the average result of 10 runs.
    \item Real-Time Video Frame Interpolation: To ensure a smooth visual output, the frame interpolation process operated within \textbf{3.97} seconds in average result of 10 runs as well, effectively reducing motion jitter and enhancing the fluidity of the avatar's movements. 
\end{itemize}

\begin{figure}[ht]
\centering
  \includegraphics[width=1\columnwidth, trim={0cm 0 0cm 0}, clip]{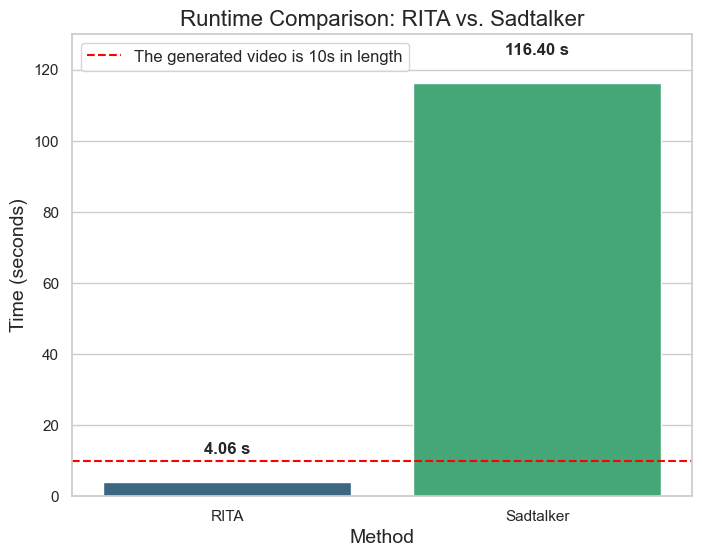}
\caption{Runtime comparison between RITA and Sadtalker. Note that generated frames by RITA can be accessed during the generation, so there is nearly no waiting time for users.}
\label{fig:timing}
\end{figure}

Figure~\ref{fig:timing} shows the total running time of RITA and SadTalker, which indicates that RITA allows avatar generation in realtime. 

\section{Conclusion}
In this study, we introduce RITA, a state-of-the-art framework for generating real-time interactive talking avatars. RITA employs Dynamic Frame Matching and Video Interpolation techniques to significantly reduce the number of frames needed during video generation, thereby enhancing the generation speed. Additionally, RITA incorporates large language models (LLMs) to build naturalistic avatar-user dialogues that elevate the interactive experience for users. Empirical results demonstrate RITA's superior performance compared to existing methods, showcasing notable improvements in generation speed, interaction quality, and user engagement.
{
    \small
    \bibliographystyle{ieeenat_fullname}
    \bibliography{main}
}


\end{document}